\documentclass[sigconf]{acmart}

\AtBeginDocument{%
  \providecommand\BibTeX{{%
    \normalfont B\kern-0.5em{\scshape i\kern-0.25em b}\kern-0.8em\TeX}}}

\setcopyright{acmcopyright}
\copyrightyear{2021}
\acmYear{2021}
\setcopyright{acmlicensed}\acmConference[CIKM '21]{Proceedings of the 30th ACM International Conference on Information and Knowledge Management}{November 1--5, 2021}{Virtual Event, QLD, Australia}
\acmBooktitle{Proceedings of the 30th ACM International Conference on Information and Knowledge Management (CIKM '21), November 1--5, 2021, Virtual Event, QLD, Australia}
\acmPrice{15.00}
\acmDOI{10.1145/3459637.3482200}
\acmISBN{978-1-4503-8446-9/21/11}

\usepackage{soul}
\usepackage{url}
\usepackage{hyperref}
\usepackage[utf8]{inputenc}
\usepackage{caption}
\usepackage{graphicx}
\usepackage{amsmath}
\usepackage{booktabs}
\usepackage{multirow}
\usepackage{tablefootnote}
\usepackage{enumitem}
\setlist[itemize]{leftmargin=*}

\usepackage{nicefrac}       % compact symbols for 1/2, etc.
\usepackage{microtype}      % microtypography

\usepackage{array}
\usepackage[mathscr]{eucal}
\usepackage{algorithm, algorithmic}
\usepackage{amsmath,amsbsy,amsfonts,amssymb,amsthm,units,bm}
\usepackage{subfiles}

\newcolumntype{C}[1]{>{\centering\let\newline\\\arraybackslash\hspace{0pt}}m{#1}}
\theoremstyle{definition}

\makeatletter
\newsavebox\myboxA
\newsavebox\myboxB
\newlength\mylenA

\newcommand*\xbar[2][0.75]{%
    \sbox{\myboxA}{$\m@th#2$}%
    \setbox\myboxB\null% Phantom box
    \ht\myboxB=\ht\myboxA%
    \dp\myboxB=\dp\myboxA%
    \wd\myboxB=#1\wd\myboxA% Scale phantom
    \sbox\myboxB{$\m@th\overline{\copy\myboxB}$}%  Overlined phantom
    \setlength\mylenA{\the\wd\myboxA}%   calc width diff
    \addtolength\mylenA{-\the\wd\myboxB}%
    \ifdim\wd\myboxB<\wd\myboxA%
       \rlap{\hskip 0.5\mylenA\usebox\myboxB}{\usebox\myboxA}%
    \else
        \hskip -0.5\mylenA\rlap{\usebox\myboxA}{\hskip 0.5\mylenA\usebox\myboxB}%
    \fi}
\makeatother

\newcommand{\tens}[1]{\boldsymbol{\mathscr{#1}}}
\newcommand{\vect}[1]{\ensuremath{\mathbf{#1}}}

\newcommand{\R}{\mathbb{R}}

\newcommand{\tA}{\tens{A}}
\newcommand{\tR}{\tens{R}}
\newcommand{\tT}{\tens{T}}
\newcommand{\tB}{\tens{B}}
\newcommand{\tX}{\tens{X}}

\newcommand{\tE}{\tens{E}}
\newcommand{\tU}{\tens{U}}
\newcommand{\tV}{\tens{V}}
\newcommand{\tS}{\tens{S}}

\newcommand{\tG}{\tens{G}}

\newcommand{\htX}{\tX_\mathscr{F}}

\newcommand{\htA}{\tA_\mathscr{F}}
\newcommand{\htR}{\tR_\mathscr{F}}

\newcommand{\A}{A}

\newcommand{\hmA}{A_\mathscr{F}}
\newcommand{\hmR}{R_\mathscr{F}}
\newcommand{\hmX}{X_\mathscr{F}}

\begin{document}
\fancyhead{}

\newcommand{\method}{{\tt Toffee}}
\newcommand{\methodr}{{\tt Toffee\textsuperscript{-}}}
\newcommand{\partitle}[1]{\smallskip \noindent \textbf{#1.}}

\title{Temporal Network Embedding via Tensor Factorization}

\author{Jing Ma$^1$, Qiuchen Zhang$^1$, Jian Lou$^{1,2}$, Li Xiong$^1$, Joyce C. Ho$^1$}
\affiliation{%
  \institution{$^1$Emory University, $^2$Xidian University}
  \country{}
}
% \affiliation{
%   \institution{$^2$Xidian University}
% }
\email{{jing.ma, qiuchen.zhang, jian.lou, lxiong, joyce.c.ho}@emory.edu, jlou@xidian.edu.cn}


\begin{abstract}
Representation learning on static graph-structured data has shown a significant impact on many real-world applications. However, less attention has been paid to the evolving nature of temporal networks, in which the edges are often changing over time. The embeddings of such temporal networks should encode both graph-structured information and the temporally evolving pattern. Existing approaches in learning temporally evolving network representations fail to capture the temporal interdependence. In this paper, we propose \method, a novel approach for temporal network representation learning based on tensor decomposition. Our method exploits the tensor-tensor product operator to encode the cross-time information, so that the periodic changes in the evolving networks can be captured. Experimental results demonstrate that \method~outperforms existing methods on multiple real-world temporal networks in generating effective embeddings for the link prediction tasks.
\end{abstract}

\keywords{Network embedding; Tensor factorization; Tensor-tensor product}

\maketitle

\section{Introduction}
Network data has gained increasing popularity due to its ubiquitous applications in multiple domains, including recommendation systems, knowledge base, and biological informatics. Learning effective embeddings over network data involves efficiently projecting the structural properties of the network data into low-dimensional feature representations which can be further utilized by many downstreaming tasks, such as node label classification, link prediction, community detection and network reconstruction.
The proposal of DeepWalk \cite{perozzi2014deepwalk} has inspired the development of many network embedding techniques \cite{tang2015line,grover2016node2vec,cao2015grarep,qiu2018network}.
However, these  techniques are all designed for static networks. While in most real-world applications, networks are usually dynamic and evolve over time. For example, in a social network, users' friend relations tend to change over time, which alters the low-dimensional feature representation. Similarly, in a co-authorship network, authors' relationship will fluctuate from time to time according to their new research projects. Therefore, it is essential to encode the temporal information into the network representation to improve the predictive power. %of the learned representations. 

To capture the evolving pattern and the temporal interaction between nodes, several temporal network embedding algorithms have been proposed.
\cite{nguyen2018continuous} proposed CTDNE, a temporal version of random walks to capture the evolving graph structure. HTNE \cite{zuo2018embedding} extended CTDNE by generating neighbors using the Hawkes process. \cite{qiu2020temporal} proposed HNIP, a temporal random walk that preserves high-order proximity and used an auto-encoder to capture the nonlinearity of the network structure.
All these neighborhood-based algorithms have an inherent disadvantage that 
their local structure inferences are heavily reliant on neighborhood relations and can overlook the global structure at large, 
which may incur poor performance especially when the graph is densely connected. This problem is further exacerbated for temporal network embedding. 
Besides, these methods lack interpretability, and require heavy tuning of hyper-parameters and model structures. 

Tensors are an efficient way to model  high-dimensional, multi-aspect data, such as spatio-temporal data \cite{zhu2016scalable, gauvin2014detecting}, where the spatial distribution at a specific timestamp will be modeled as a frontal slice of the tensor.
Built on the success of the matrix factorization-based network embedding methods \cite{cao2015grarep, qiu2018network} in revealing useful global structure information of the network, tensors are able to capture both the global network structure and the temporal node interactions at the same time. Due to the inefficacy of traditional tensor factorization approaches such as CANDECOMP/PARAFAC (CP) and Tucker decomposition %in modeling the collective learning paradigm which 
to take advantage of the slice-wise correlations, RESCAL \cite{nickel2011three} was proposed %as an efficient collective learning algorithm based on tensor factorization, which shows superior performance in 
to learn
representations of multi-relational data. Nonetheless, RESCAL is still insufficient in capturing the temporal evolution by a single embedding matrix.

In this paper, we propose a \textbf{T}ens\textbf{o}r \textbf{f}actorization algorithm \textbf{f}or t\textbf{e}mporal network \textbf{e}mbedding (\method), which models the temporal network structure as a three-way tensor, and learns unique representations for the temporal network through a novel tensor factorization algorithm. \method~is superior to RESCAL by capturing the cross-time relation. This is empowered by the tensor-tensor product (t-product) \cite{kilmer2011factorization, zhang2014novel}, which has demonstrated great potential in image/video domain by  exploiting the circular convolution operator along the temporal dimension.

We briefly summarize our contributions as follows. 
1) We propose \method, a new tensor factorization model that incorporates the t-product for temporal network embedding. 
\method~ manifests superior capability in capturing both the global structure information and the temporal interactions between nodes. 
2) \method~is able to extract unique representations for asymmetric relationships between nodes (both directed and indirected).
3) Experiments on multiple real-world network datasets show that \method~is capable of generating embeddings with better predictive power compared to state-of-the-art approaches in temporal network embedding. %evaluated on 

\section{Preliminaries and Notations}
In this section, we introduce the frequently used definitions and notations in this paper. 
We use 
$\A$ to denote a matrix, $\tA$ to denote a tensor, and $\tA(i,:,:)$, $\tA(:,i,:)$, $\tA(:,:,i)$ denote the horizontal, lateral and frontal slice of tensor $\tA$, respectively. We use $\tA^{(i)}$ to represent $\tA(:,:,i)$. We denote $\htA = {\tt fft}(\tA,[\hspace{1mm}],d)$ as the tensor after fast Fourier transform (fft) along the $d$-th mode, and $\htA$ can be transformed back to $\tA$ through the inverse fft $\tA = {\tt ifft}(\htA,[~],d)$.

\begin{definition}
(\textit{Temporal Network}) A temporal network $\tG$ is denoted as $\tG =(\tV, \tE, \tT)$, where $\tV$ is a set of nodes, $\tE$ is the set of edges, and $\tT$ is the sequence of timestamps. Given $t\in\tT$, we will have a sequence of network snapshots $\tG_t=(\tV_t, \tE_t)$ at time $t$, where $\tV_t$ are a set of nodes at time $t$, and $\tE_t$ are the set of interactions between $\tV_t$.
\end{definition}

\begin{definition} \textbf{(t-product \cite{kilmer2011factorization,kilmer2013third})}
\label{def.t-product}
  The t-product between tensor $\tA\in\mathbb{R}^{n_1\times n_2\times n_3}$ and $\tB\in\mathbb{R}^{n_2\times n_4\times n_3}$ is defined as $ \tA * \tB = \tens{C}\in\mathbb{R}^{n_1\times n_4\times n_3}$ with the $(i,j)$-th tube $\mathring{\vect{c}}_{ij}$ of $\tens{C}$ computed as
  \begin{small}
  \begin{equation}
  \label{Eq.fiber_prod}
    \mathring{\vect{c}}_{ij} = \tens{C}(i,j,:) = \sum _{k=1}^{n_2}\tA(i,k,:)*\tB(k,j,:),
  \end{equation}
  \end{small}
  where $*$ denotes the circular convolution \cite{rader1968discrete} between two tubes of the same size.
\end{definition}
% From Def. \ref{def.t-product}, the t-product is analogous to the matrix product in a way that it replaces the multiplication between scalars (i.e. the $\cdot$ in $C_{ij} = \sum _{k=1}^{n_2} A(i,k)\cdot B(k,j)$) to the circular convolution between fibers (i.e. the $*$ in Eq.(\ref{Eq.fiber_prod})). 

\begin{definition} \textbf{(Block Diagonal Matrix \cite{zhang2014novel})}
\label{def.block_diag_matrix}
The block diagonal matrix $\hmA$ is computed as
  \begin{small}
  \begin{equation}
  \label{Eq.fourier.matrix}
    \begin{aligned}
    \hmA := &{\tt blockdiag}(\htA)\\
     := &\left[\begin{array}{cccc}\hmA^{(1)}& & & \\
     & \hmA^{(2)} & & \\
     & &\ddots & \\
      & & & \hmA^{(n_3)}\end{array} \right]
      \in \mathbb{C}^{n_1n_3 \times n_2n_3},
    \end{aligned}
\end{equation}
\end{small}
where $\htA = {\tt fft}(\tA,[\hspace{1mm}],3)$ is the tensor after fft along the third mode, and $\hmA^{(i)} = \htA^{(i)}$, for $i=1$ to $n_3$.
\end{definition}

\begin{definition}
\textbf{(RESCAL \cite{nickel2011three})}
RESCAL is a tensor factorization algorithm which 
forms the multi-relational data as a tensor $\tX: \R^{n\times n \times m}$, and employs the rank-$r$ decomposition for each tensor slice 
as $\tX^{(k)} \approx \A\tR^{(k)}\A^T$, for $k=1, \dots, m$.
where $\A$ is an $n\times r$ matrix representing the latent components of each entities and $\tR^{(k)}$ is an $r\times r$ matrix representing the interactions of the latent components of $k$-th relation.
\end{definition}

\begin{definition}
\textbf{(t-SVD \cite{kilmer2011factorization,kilmer2013third, zhang2014novel})}
t-SVD factorizes a tensor $\tX: \R^{n_1\times n_2 \times n_3}$ as $\tX=\tU*\tS*\tV$, where $*$ denotes the t-product, $\tU$ and $\tV$ are orthogonal tensors of size $n_1\times n_1 \times n_3$ and $n_2\times n_2 \times n_3$ respectively. $\tS$ is an f-diagonal tensor of size $n_1\times n_2 \times n_3$ where each frontal slice is diagonal.
\end{definition}

\section{Proposed Method}

In this section, we formally define the temporal network embedding algorithm by formulating the the objective function as a tensor reconstruction problem. 

\begin{figure}
\centering
\includegraphics[width=0.4\textwidth]{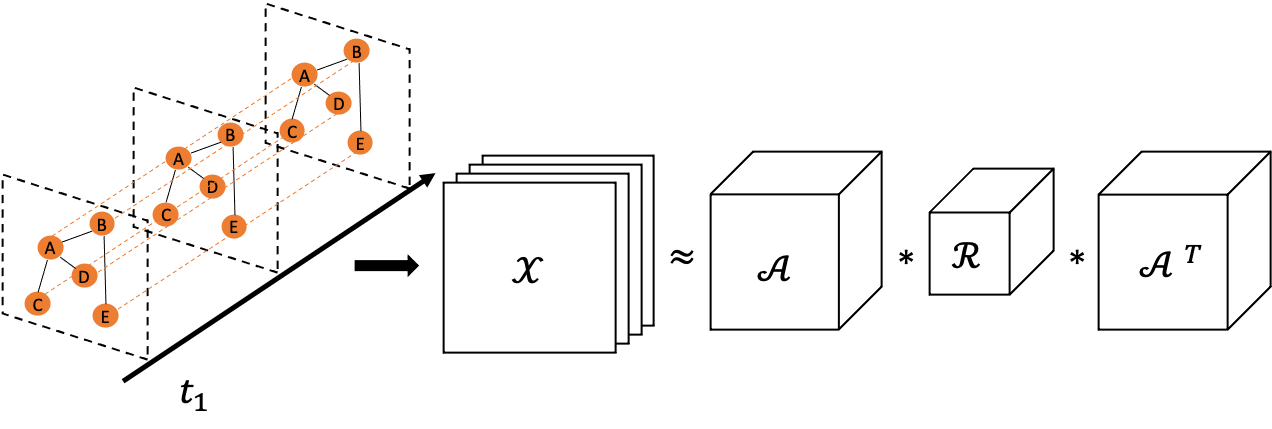}
\vspace{-1em}
\caption{Algorithm Overview}
\label{fig:algo}
\vspace{-1em}
\end{figure}
 
\subsection{Problem Formulation}
Given a temporal network consisting of a sequence of network snapshots $\tG=\{\tG_1,\dots,\tG_T \}$, we construct a temporal adjacency tensor $\tX: \R^{n\times n \times T}$, each of the frontal slice of the tensor $\tX(:,:,t)$ is set to the adjacency matrix of the network snapshot at timestamp $t=\{1, \dots, T\}$ (Fig. \ref{fig:algo}). In this way, we extend the modeling of the network structure to a three-way tensor with the temporal evolution explicitly spanned and accommodated along the third dimension. Our key intuition is that the decomposition structure of RESCAL, especially its request of a single matrix $\A$ to be shared by all timestamps, is not sufficient to fully capture the fruitful temporally evolving information. In contrast, we introduce $T$ matrices and collect them into the $n\times r\times T$ tensor $\tA$, so that each of the frontal slice $\tA(:,:,t)$ models the latent representation of each node at time $t$. In addition, we propose to capture the cross-slice intercorrelation among the T matrices of $\tA$, together with the same $r\times r\times T$ tensor $\tR$ as RESCAL, by the circular convolution-based t-product. As a result, we formulate the objective function of \method~ as follows\footnote{$(\cdot)^{\top}$ denotes the conjugate transpose in this paper.},

\begin{small}
\begin{equation}
\label{eq.obj1}
  \arg\min_{\tA,\tR} \frac{1}{2}\|\tX - \tA*\tR*\tA^{\top}\|_F^2 + \frac{\lambda_A}{2} \|\tA\|_F^2 + \frac{\lambda_R}{2} \|\tR\|_F^2,
\end{equation}
\end{small}
where the regularizations with parameters $\lambda_A$ and $\lambda_R$ are introduced on $\tA$ and $\tR$ to prevent overfitting. \method~ can also support other regularizations but we omit more complicated choices for illustrative purposes.

The optimization of the \method~ model takes advantage of the well-known relationship between the convolution operation and the Fourier transformation \cite{kolba1977prime, kilmer2013third}, which converts the problem to Fourier domain and is parallelizable among $T$ slices. In detail, we first transform the temporal tensor $\tX$ to the Fourier domain along the temporal mode by $\htX = {\tt fft}(\tX,[\hspace{1mm}],3)$, and then perform the rank-$r$ factorization on the block diagonal matrix $\hmX:={\tt blockdiag}(\htX)$ into the block diagonal matrices of $\tA$ and $\tR$.
This can be further decomposed into $T$ independent optimization problems by decomposing each block of the diagonal $\hmX^{(k)}$ with the $k\in [t]$-th being,
\begin{small}
\begin{equation*}
  \begin{split}
      \arg\min_{\hmA^{(k)},\hmR^{(k)}} & \frac{1}{2T}\|\hmX^{(k)} - \hmA^{(k)}\cdot\hmR^{(k)}\cdot(\hmA^{(k)})^{\top}\|_F^2 \\
      & + \frac{\lambda_A}{2T} \|\hmA^{(k)}\|_F^2 + \frac{\lambda_R}{2T} \|\hmR^{(k)}\|_F^2,
  \end{split}
\end{equation*}
\end{small}
where $\cdot$ denotes the matrix multiplication, $\hmA^{(k)}$ and $\hmR^{(k)}$ can be seen as each block of the block diagonal matrices $\hmA$ and $\hmR$. We then optimize them alternatively for each of the $T$-blocks in parallel.

\subsection{Optimization}

\partitle{Update $\hmA^{(k)}$}
To derive the update rule of $\hmA^{(k)}$, we extend the ASALSAN algorithm \cite{bader2007temporal} to Fourier domain, where we first fix $\hmR^{(k)}$ and
convert the objective function regarding $\hmA^{(k)}$ as
\begin{small}
\begin{equation*}
  \arg\min_{\hmA^{(k)}} \frac{1}{2r}\|\xbar{\hmX^{(k)}} - \hmA^{(k)}\cdot\xbar{\hmR^{(k)}}\cdot\left(I \otimes (\xbar{\hmA^{(k)})}^{\top}\right)\|_F^2 + \frac{\lambda_A}{2r} (\|\hmA^{(k)}\|_F^2),
\end{equation*}
\end{small}
where $\xbar{\hmX^{(k)}}$, $\xbar{\hmR^{(k)}}$ and $\xbar{\hmA^{(k)}}$ denote $[\hmX^{(k)} ~~ (\hmX^{(k)})^{\top}]$, $[\hmR^{(k)} ~~ (\hmR^{(k)})^{\top}]$ and the right $\hmA^{(k)}$, respectively, $I$ is the identity matrix with size $n\times n$, and $\otimes$ denotes the \textit{kronecker product}. We optimize the above objective function by keeping $\xbar{\hmA^{(k)}}$ as constant, and update only the left $\hmA^{(k)}$. Taking the derivative with respect to $\hmA^{(k)}$, we get the following closed-form solution
\begin{small}
\begin{equation*}
  \label{eq.update.Akfft}
  \begin{split}
  \hmA^{(k)} = &\Big{(}\hmX^{(k)}\hmA^{(k)}(\hmR^{(k)})^{\top} + (\hmX^{(k)})^{\top}\hmA^{(k)}\hmR^{(k)}\Big{)} \\
  & \cdot \Big{(}\hmR^{(k)}(\hmA^{(k)})^{\top}\hmA^{(k)}(\hmR^{(k)})^{\top}+(\hmR^{(k)})^{\top}(\hmA^{(k)})^{\top}\hmA^{(k)}\hmR^{(k)}+\lambda_A I\Big{)} ^{-1}.
  \end{split}
\end{equation*}
\end{small}

\vspace{-0.5em}

\partitle{Update $\hmR^{(k)}$}
The objective function after fixing $\hmA^{(k)}$ and vectorizing $\hmX^{(k)}$ and $\hmR^{(k)}$ is as follows
\begin{small}
\begin{equation*}
       \arg\min_{\hmR^{(k)}} \frac{1}{2r}\|{\tt vec}(\hmX^{(k)}) - (\hmA^{(k)}\otimes\hmA^{(k)})\cdot{\tt vec}(\hmR^{(k)})\|_F^2 + \frac{\lambda_R}{2r} (\|{\tt vec}(\hmR^{(k)})\|_2^2).
\end{equation*}
\end{small}
\vspace{-0.3em}
Taking the derivative with respect to ${\tt vec}(\hmR^{(k)})$ and setting it to zero, we obtain the update for $\hmR^{(k)}$ as follows
\begin{small}
\begin{equation*}
  \label{eq.update.Rkfft}
  \begin{split}
  \hmR^{(k)} = & ((\hmA^{(k)}\otimes\hmA^{(k)})^{\top}(\hmA^{(k)}\otimes\hmA^{(k)})+\lambda_R I)^{-1}\\
  & \cdot (\hmA^{(k)}\otimes\hmA^{(k)})\cdot {\tt vec}(\hmX^{(k)}).
  \end{split}
\end{equation*}
\end{small}
$\hmA^{(k)}$ and $\hmR^{(k)}$ are alternatively optimized until the objective converges. We then reconstruct the resulting tensor $\tA,\tR$ by converting the block diagonal matrices back to the temporal domain using the inverse fft: $\tA = {\tt ifft}(\htA,[~],3)$, $\tR = {\tt ifft}(\htR,[~],3)$.

\partitle{Embedding Design} Equipped with the factorization output, we consider the design of the embedding, i.e., compose a unique embedding for each node consisting all the temporal information. Our intuition is that such embedding should involve as much cross-timestamp interrelation as possible through the t-product as $\tA(i,:,:)*\tR$, where $\tA(i,:,:)$ involves the factorized temporal information of all time for node $i$. This yields a tensor of size $1\times r\times T$, which can be squeezed into an $r\times T$ matrix. We then generate the final embedding through summing along the temporal mode ($2$-nd mode) to get an $r$-dimension vector embedding for node $i$:
\begin{equation}
\label{eq.emb}
\mathbf{emb_i} = \sum_{t=1}^T{\tt squeeze}(\tA(i,:,:)*\tR).
\end{equation}

\subsection{Complexity Analysis}
For a temporal adjacency tensor $\tX: \R^{n\times n \times T}$, FFT and inverse FFT operations takes $O(N^2T \log(T))$ for each iteration. The computational cost of $T$ $n\times n$ matrices is dominated by the computation of $\hmR^{(k)}(\hmA^{(k)})^{\top}\hmA^{(k)}(\hmR^{(k)})^{\top}$ and $(\hmR^{(k)})^{\top}(\hmA^{(k)})^{\top}\hmA^{(k)}\hmR^{(k)}$ which has the complexity of $O(2(2r^2N+r^3))$, where $r$ is the rank and is usually a small value. The total computational complexity is summarized as $O(N^2T \log(T))+O(2T(2r^2N+r^3))$. The FFT operation requires forming the tensor $\tX$ and the output tensors $\tA$ and $\tR$, which has the storage complexity of $O(n^2T+nrT+r^2T)$.

\section{Experiments}
Similar to \cite{nguyen2018continuous}, we evaluate the network embedding generated by \method~with the link prediction task. In the experiments, we answer two important questions:

\textbf{RQ1:} How does \method~compare with other state-of-the-art temporal network embedding algorithms?

\textbf{RQ2:} Does the key design in \method~help in achieving better performance?

\begin{small}
\begin{table}
  \centering
  \begin{tabular}{ l|c|c|c|c }
    \hline
    \textbf{Dataset}&\textbf{\# Nodes} & \textbf{\# Edges} & $\bar{d}$ &\textbf{Timespan} \\
    \hline
    fb-forum & 899 & 34K & 74 & 166\\
    email-Eu-core\tablefootnote{This dataset is from: https://snap.stanford.edu/data/email-Eu-core.html} & 986 & 26K & 25.44 & 526\\
    ia-primary-school-proximity & 242 & 126K & 1K & 3100\\
    ia-contacts-hypertext2009 & 113 & 21K & 368 & 4\\
    ia-workplace-contacts & 92 & 10K & 213 & 7104\\
    aves-wildbird-network & 202 & 11.9K & 117 & 2884\\
     \hline
\end{tabular}
\caption{Dataset Statistics}
\label{table:dataset}
\vspace{-2em}
\end{table}
\end{small}
% \vspace{-1em}

\begin{small}
\begin{table*}[h]
  \centering
  \begin{tabular}{ l|cccccc|cc }
  \toprule
    % \hline
    \textbf{Dataset}& \textbf{HTNE} & \textbf{CTDNE} & \textbf{HNIP} & \textbf{TNE} & \textbf{t-SVD} & \textbf{RESCAL} & \textbf{\methodr} & \textbf{\method}\\
    \midrule
    fb-forum & 0.8615 & 0.8567 &0.8091 & 0.6493&0.8885	&0.8010 & 0.9280 & \textbf{0.9420}\\
    email-Eu-core & 0.8737& 0.8320& 0.8312 & 0.6615&0.8684	&0.8812 &0.8912 & \textbf{0.9010}\\
    ia-primary-school-proximity & 0.7957& 0.7806 & 0.7957 & 0.6043&0.8213	&0.8102 & 0.8162 & \textbf{0.8218}\\
    ia-contacts-hypertext2009 & 0.6956& 0.5831& 0.5832 & 0.6526&0.7669	&0.7788 & 0.7687 & \textbf{0.9661}\\
    ia-workplace-contacts & 0.7834& 0.6653& 0.7834 & 0.6082&0.7835	&0.7146 & 0.7913 & \textbf{0.8366}\\
    aves-wildbird-network & 0.8869& 0.8901& 0.8066 & 0.7018&0.9778	&0.9202& 0.9851&\textbf{0.9936}\\
     \bottomrule
\end{tabular}
\caption{Link Prediction Results (Micro-F1). The best scores for each operators on each dataset are underlined, the best-performance operator algorithm is in bold.}
\label{table:result}
\vspace{-1em}
\end{table*}
\end{small}

\subsection{Baselines}
We compare \method~with the following state-of-the-art temporal network embedding baselines

\begin{itemize}

\item{\textbf{TNE} \cite{zhu2016scalable}}.
A matrix factorization-based temporal network embedding algorithm that jointly factorizes the temporal adjacency matrices with a uniqueness penalty term on the embeddings.

\item{\textbf{CTDNE} \cite{nguyen2018continuous}}. 
A temporal network embedding algorithm that exploits the time-constrained temporal random walk using an extension to the node2vec \cite{grover2016node2vec} algorithm via the skip-gram model.

\item{\textbf{HNIP} \cite{qiu2020temporal}}.
A temporal network embedding algorithm that employs the high-order non-linear information as an extension to the auto-encoder-based network embedding algorithms.

\item{\textbf{HTNE} \cite{zuo2018embedding}}.
A temporal network embedding algorithm that uses a Hawkes process to capture the influence of the historical neighbors on the current neighbor.
\end{itemize}

To investigate the benefits of the key components, we also compare our proposed \method~with \textbf{RESCAL} \cite{nickel2011three} and \textbf{t-SVD} \cite{kilmer2011factorization, zhang2014novel} which differs from \method~in that RESCAL uses a single matrix as representation of each entity (node), and t-SVD solves the truncated SVD in the Fourier domain, which will result in an f-diagonal middle tensor $\tS$. We also compare with \methodr, which is \method~without regularizations.

\begin{figure}[b]
\centering
\vspace{-0.8em}
\includegraphics[width=0.4\textwidth]{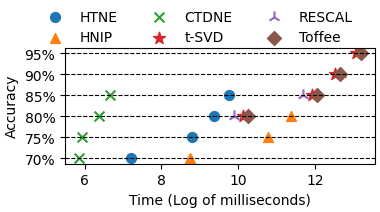}
\caption{The time it takes for each method to reach different levels of accuracy for aves-wildbird-network dataset.}
\label{fig:time}
\vspace{-0.5em}
\end{figure}

\subsection{Experiment Setup}

Experiments are conducted on six temporal network datasets from the Network Repository \cite{nr} which are densely connected (with high average degrees $\bar{d}$) to verify the effectiveness of \method~in capturing global structure information. The statistics of the temporal networks are summarized in Table \ref{table:dataset}.

The regularization parameters $\lambda_A$ and $\lambda_R$ in \method~is determined through grid-search among $\{1^{-4}, 1^{-3}, 1^{-2}, 1^{-1}\}$.
To fairly compare with RESCAL and t-SVD, we use the same embedding construction method as \method~in eq.(\ref{eq.emb}) to encode sufficient temporal information (for RESCAL we use matrix product instead of t-product). 
For CTDNE, HNIP, and HTNE, we adopt the same hyper-parameter settings as in paper \cite{qiu2020temporal}. For TNE, we take the embeddings of last timestamp for evaluations.
For all the tested algorithms, the embedding dimension is set to 128 (ia-contacts-hypertext2009 and ia-workplace-contacts have embedding dimension as 64 since the number of nodes are less than 128). 

For the link prediction task, we adopt the same experiment settings as in \cite{nguyen2018continuous} that we select the first 75\% of the temporal links for learning the embeddings, and train a logistic regression classifier on the rest 25\% of the temporal links as positive examples, combined with the same number of negative examples generated by random sampling from the non-existing links. The links between two nodes are defined as edge representations and are calculated with four different operators defined in \cite{grover2016node2vec}. We report the average Micro-F1 score of 10 replications based on 10 different seeds of initializations.

\subsection{Link Prediction}

Similar to \cite{nguyen2018continuous,qiu2018network,zuo2018embedding}, we evaluate the effectiveness of the \method~extracted embeddings on the link prediction task. The aim of link prediction task is to determine whether there will be an edge between two nodes in the future based on the embeddings learned from the past to measure the prediction ability of the embeddings. 

Table \ref{table:result} shows the Micro-F1 score for the link prediction task with edge representation computed by multiple operators. 
First of all, results demonstrate that \method~significantly outperforms other baselines considering the best-performance operator (\textbf{RQ1}). This is due to the fact that \method~can better exploit the global network structure compared with the baselines which infer the network embeddings based on neighborhood information. 
Specifically, all tensor based algorithms including t-SVD, RESCAL, and \method~have better performance compared with the neighborhood-based algorithms CTDNE, HNIP, and HTNE considering the best-achievable operator, which also verifies that tensor based methods can better capture the global structure information. Those tensor-based methods also beat the matrix-based method TNE, indicating the superiority of tensors in jointly modeling the temporal and structural information between nodes.
In addition, it is also worth noticing that even with no regularizations on tensor $\tA$ and $\tR$, \methodr~can outperform the baselines. This means that \method~can help reducing the effort of hyper-parameter tuning.

Fig. \ref{fig:time} shows that \method~achieves significantly better accuracy with more time cost compared with HTNE, HNIP, CTDNE -- illustrating that in order to get more accurate network representations, we have to sacrifice some time cost. However, compared with tensor-based methods, t-SVD and RESCAL, which offer superior performance than HTNE, HNIP, CTDNE, \method~is able to achieve better accuracy with little trade-off in terms of computational cost.

Furthermore, we observe that \method~outperforms both t-SVD and RESCAL (\textbf{RQ2}).
The performance gain over RESCAL verifies that \method~benefits from the tensor product which will help better capture the temporal correlation by the circular convolution operator.
Notice that without regularization, \methodr~resembles the results of t-SVD with slight higher accuracy. This is because each slice in tensor $\tR$ is able to capture the directional information between the latent groups of each node, which is not efficiently encoded in the results of t-SVD with information concentrated only on the diagonals. In addition to this, \method~largely outperforms t-SVD thanks to the regularization terms on tensors $\tA$ and $\tR$.

\section{Conclusion}
In this paper, we have proposed \method, an advanced tensor decomposition algorithm for temporal network embedding. The exploitation of t-product enables \method~to better capture the cross-time information among different temporal slices of the temporal network. We demonstrate the effectiveness of \method~with the temporal link prediction task with multiple real-world datasets. Experiment results show significant improvement over state-of-the-art temporal network embedding algorithms. Future works include involving the attribute information and further demonstrating the effectiveness of \method~on other downstreaming tasks.

\begin{acks}
This work was supported by the National Science Foundation under award IIS-\#1838200 and CNS-1952192, National Institute of Health (NIH) under award number R01LM013323, K01LM012924 and R01GM118609, CTSA Award UL1TR002378, and Cisco Research University Award \#2738379.
\end{acks}

\newpage
\bibliographystyle{ACM-Reference-Format}
\bibliography{sample-base}

\end{document}